\documentclass[10pt,twocolumn,letterpaper]{article}

\PassOptionsToPackage{table}{xcolor}

\usepackage{cvpr}              

\usepackage{algorithm}
\usepackage[noend]{algpseudocode}

\definecolor{cvprblue}{rgb}{0.21,0.49,0.74}
\usepackage[pagebackref,breaklinks,colorlinks,allcolors=cvprblue]{hyperref}
\usepackage{multirow}
\usepackage{float}
\usepackage{adjustbox}
\definecolor{goldcolor}{RGB}{255,215,0}
\definecolor{silvercolor}{RGB}{192,192,192}
\definecolor{bronzecolor}{RGB}{205,127,50}

\definecolor{bestColor}{RGB}{219,238,255}
\definecolor{worstColor}{RGB}{255,202,189}

\newcommand{\BestColor}{bestColor!55!white}  
\newcommand{\WorstColor}{worstColor!35!white}
\newcommand{\best}[1]{\cellcolor{\BestColor}#1}
\newcommand{\worst}[1]{\cellcolor{\WorstColor}#1}
\newcommand{\BestavgColor}{bestColor}  
\newcommand{\WorstavgColor}{worstColor}
\newcommand{\bestavg}[1]{\cellcolor{\BestavgColor}#1}
\newcommand{\worstavg}[1]{\cellcolor{\WorstavgColor}#1}

\def\confName{CVPR}
\def\confYear{2026}

\title{ChArtist: Generating Pictorial Charts with Unified Spatial and Subject Control}

\author{Shishi Xiao\footnotemark[1]\\
Brown University\\
{\tt\small shishi\_xiao@brown.edu}
\and
Tongyu Zhou\\
Adobe Research\\
{\tt\small tongyuz@adobe.com}
\and
David H. Laidlaw\\
Brown University\\
{\tt\small dhl@cs.brown.edu}
\and
Gromit Yeuk-Yin Chan\\
Adobe Research\\
{\tt\small ychan@adobe.com}
}

\begin{document}
\twocolumn[{
\renewcommand\twocolumn[1][]{#1}
\maketitle

\begin{center}
\centering
  \centering
  \vspace{-1em}
  \includegraphics[width=0.98\linewidth]{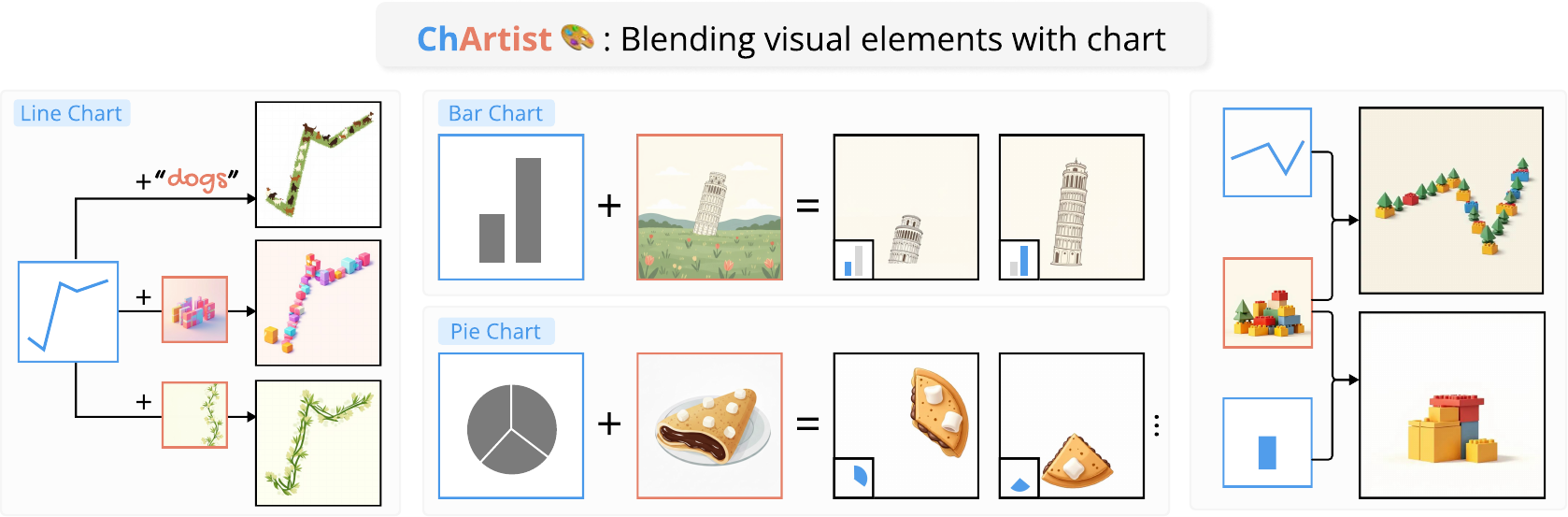}
  \captionof{figure}{
    Illustrations of pictorial charts generated by ChArtist. We convert chart primitives such as bars, lines, and segments into vivid visual elements. With user-provided text or reference images, ChArtist combines spatial and subject control to maintain data fidelity while achieving visual consistency.
  }
  \label{fig:Teaser}
\end{center}
}]
\footnotetext[1]{Work done during internship at Adobe Research.}
\begin{abstract}
A pictorial chart is an effective medium for visual storytelling, seamlessly integrating visual elements with data charts. However, creating such images is challenging because the flexibility of visual elements often conflicts with the rigidity of chart structures. This process thus requires a creative deformation that maintains both data faithfulness and visual aesthetics. Current methods that extract dense structural cues from natural images (e.g., edge or depth maps) are ill-suited as conditioning signals for pictorial chart generation.
We present ChArtist, a domain-specific diffusion model for generating pictorial charts automatically, offering two distinct types of control: 1) spatial control that aligns well with the chart structure, and 2) subject-driven control that respects the visual characteristics of a reference image. To achieve this, we introduce a skeleton-based spatial control representation. This representation encodes only the data-encoding information of the chart, allowing for the easy incorporation of reference visuals without a rigid outline constraint. We implement our method based on the Diffusion Transformer (DiT) and leverage an adaptive position encoding mechanism to manage these two controls.
We further introduce Spatially Gated Attention to modulate the interaction between spatial control and subject control.
To support the fine-tuning of pre-trained models for this task, we created a large-scale dataset of 30,000 triplets (skeleton, reference image, pictorial chart). We also propose a unified data accuracy metric to evaluate the data faithfulness of the generated charts.
We believe this work demonstrates that current generative models can achieve data-driven visual storytelling by moving beyond general-purpose conditions to task-specific representations.
Project page: \url{https://chartist-ai.github.io/}.

\end{abstract}    
\section{Introduction}
\label{sec:intro}

Pictorial charts that embed semantic data directly into chart structures provide a rich and engaging way to tell visual stories~\cite{park2019graphoto, coelho2020infomages, xiao2023let, wu2023viz2viz}. These visualizations captivate viewers by blending representative imagery with data, converting information into experiences that enhance both understanding and long-term recall~\cite{borgo2012empirical,haroz2015isotype}.

However, creating such a pictorial chart from scratch is highly challenging, involving both creativity and design skills. The integration of malleable visual imagery with the rigid shape of a standard chart requires a careful balance of data faithfulness and visual aesthetics. 
Some examples of this are shown in Fig.~\ref{fig:Teaser}. 
For example, to create a line chart about dogs, the designer may first query for and identify an image that aligns with the overall trend of the line chart, then overlay the chart on top~\cite{park2019graphoto, coelho2020infomages}. Alternatively, they could also use collage tools to fill in data points on the line chart with custom glyphs~\cite{shi2022supporting, cui2021mixed}; this is operationally simpler at the cost of the pictorial chart's expressiveness.
While recent authoring tools have been developed to assist in pictorial chart creation, these workflows still require substantial manual refinement and lack effective quantitative evaluation of data faithfulness~\cite{xiao2023let, wu2023viz2viz}.


In this paper, we propose an end-to-end pipeline for pictorial chart generation that supports two types of control: 1) \textit{spatial} and 2) \textit{subject-driven}. These controls were inspired by observations of real design workflows: users either adopted a ``data-first'' approach---finalizing the chart first and searching for compatible images afterwards---or a ``visual-first'' approach---selecting an appealing image and deforming it to fit the chart structure.

Existing spatial-control methods typically rely on image-based conditions such as Canny edges~\cite{isola2017image, zhang2023adding, mou2024t2i, tan2025ominicontrol, luo2024readout}, which are designed for natural images that contain far greater structural complexity than charts.
To tackle this problem, we propose a new control representation for data charts.
By focusing solely on the data-encoding dimension, each chart is condensed to a skeleton representation that enables semantic and stylistic adaptability during generation.
As shown in Fig.~\ref{fig:sk_method}, this skeleton encodes semantically meaningful visual properties such as bar heights, line trends, and pie-slice angles.

Our approach builds upon FLUX~\cite{labs2025flux1kontextflowmatching}, a pretrained Diffusion Transformer (DiT) model, and trains two task-specific LoRA modules: $LoRA_{S}$, which enforces spatial control from the chart skeleton, and $LoRA_{R}$, which injects subject control from reference images. These two controls can be applied independently or jointly to support diverse design workflows, as illustrated in Fig.~\ref{fig:pipe} B.

However, composing multiple LoRAs in parallel introduces substantial cross-condition interference, a common issue in multi-control setups, yet particularly harmful for charts, where any subject-induced distortion may directly break data faithfulness. To address this, we propose Spatially-Gated Attention, which sequentially modulates the interaction between spatial and subject control, establishing an explicit dependency between them during inference and substantially improving spatial fidelity and visual consistency under harmonious dual control.

To validate the effectiveness of our approach, we construct \textsc{ChArtist-30K}, a high-quality synthetic dataset of 30,000 skeleton–reference–pictorial chart triplets, generated through two specialized pipelines tailored to distinct chart topologies.
In addition, we introduce a unified data accuracy metric to quantitatively evaluate different forms of pictorial charts.
We hope that this dataset and evaluation framework will foster future research on data-driven storytelling with generative models.
Our evaluations show that ChArtist achieves robust balance between data faithfulness and visual expressiveness.
Our key contributions are summarized as follows:
\begin{itemize}
    \item A chart-specific control representation that is minimal, precise, and flexible.
    \item An end-to-end framework for pictorial chart generation with both spatial and subject control.
    \item A Spatially-Gated Attention mechanism to mitigate cross-condition interference.
    \item A \textsc{ChArtist-30K} dataset and a unified data-accuracy metric for pictorial charts.
\end{itemize}

\section{Related Works}
\label{sec: related}

\subsection{Control Representations for Generative Models}
Current generative models have established a series of control representations that can be broadly categorized by their levels of granularity.
On one hand, dense representations such as Canny edges, depth maps, and semantic segmentation maps~\cite{isola2017image, zhang2023adding, mou2024t2i, tan2025ominicontrol, luo2024readout} can provide strict guidance and are effective for tasks such as photorealistic generation.
While powerful, these dense representations assume pixel-perfect correspondence between the conditions and the generated image.
This rigidity restricts the range of visual deformations needed for expressive tasks.
Recent work, such as LooseControl~\cite{bhat2024loosecontrol}, has identified this limitation and proposed a looser adherence to depth conditions.
More relevant to our work are abstract, sparse representations, including sketches~\cite{voynov2023sketch, zhu2016generative, koley2023picture}, bounding boxes~\cite{li2023gligen, Zhao_2019_CVPR, zheng2023layoutdiffusion}, and human pose keypoints~\cite{ma2017pose, ju2023humansd, chan2019everybody}.
These representations allow for structural control while preserving stylistic creativity. 
However, directly applying them to pictorial chart generation remains challenging: sketches focus on overall shape instead of data-encoding dimension, and pose keypoints rely on a fixed set of joints, making them unsuitable for generalizing across chart types with varying structures. 
In response, we introduce a control representation specifically for charts, enabling precise data information while generalizing consistently across different chart topologies.

\subsection{Controllable generation}
\noindent \textbf{Spatially Aligned Generation}
It typically employs image-conditioned controls to enforce structural constraints.
Some approaches inject structural features into frozen diffusion backbones through task-specific branches or adapters ~\cite{zhang2023adding, mou2024t2i, liu2024smartcontrol}.
Recent studies move toward unified and multimodal conditioning frameworks~\cite{zhao2023uni, huang2023composer, kim2023diffblender, le2025one}, which encode diverse spatial priors into shared representations, enabling cross-condition generalization.

\noindent \textbf{Subject Driven Generation}
Subject-driven generation aims to preserve the identity of a reference subject while generating it across diverse visual contexts.
Pioneering works rely on subject-specific optimization for each new concept, whether by optimizing model weights~\cite{ruiz2023dreambooth, kumari2023multi}, text embeddings~\cite{wei2023elite, sohn2023styledrop}, or modulation signals in diffusion transformers~\cite{garibi2025tokenverse, zhong2025mod}. 
Some methods achieve high-fidelity subject preservation by leveraging the in-context generation capability of pretrained diffusion models, either via lightweight LoRA tuning or by direct inpainting with side-by-side concatenated images~\cite{huang2024context, shin2025large, gu2024analogist}.

\noindent \textbf{Unified Generation}
Recent works aim to unify spatial and subject control within a single framework.
Luo et al.~\cite{luo2024readoutguidance} propose parameter-efficient readout heads that extract intermediate features and project them into the target domain, which are used to guide latent optimization during sampling.
UniCombine~\cite{wang2025unicombine} and OmniControl~\cite{tan2025ominicontrol} leverage unified sequence processing for DiT, allowing flexible token interactions.
Building on this line of research, we extend unified controllable generation to jointly model spatial constraints and reference-driven visual consistency in pictorial chart synthesis.

\subsection{Visual Blending}
Visual blending, the process of integrating multiple visual concepts into a coherent and interpretable composition, is an effective strategy for visual communication. It has been widely explored in various forms, such as creating illusion art~\cite{geng2024visual, burgert2024diffusion}, hiding QR codes in images~\cite{zhang2015aesthetic, xu2021art}, and semantic text stylization~\cite{iluz2023word, yang2019tet, xiao2024typedance, tendulkar2019trick, yang2018context}.
A key challenge across these tasks is preserving both structural fidelity and visual expressiveness.
To tackle this, earlier work often relied on retrieval-based composition~\cite{tendulkar2019trick} to match shape constraints or on style transfer techniques to apply texture to a given contour~\cite{fish2020sketchpatch, azadi2018multi}. More recent approaches, particularly in semantic typography, optimize the outline itself to incorporate new concepts by leveraging the rich natural image priors from pretrained diffusion models~\cite{iluz2023word}.
Our work extends visual blending to data charts, a domain that demands even stricter structural guidance due to the need to encode precise numerical data. We aim to achieve this through an end-to-end generative pipeline, rather than manual authoring tools~\cite{coelho2020infomages, xiao2023let, zhang2020dataquilt}, and we introduce a unified metric for evaluating data accuracy.

\section{\textsc{ChArtist-30K} Dataset}
\label{sec: dataset}

We construct a large-scale synthetic dataset including 30,000 triplets for pictorial chart generation.
Each sample is a $(S, R, P)$ triplet where $S$ is the chart skeleton image, $R$ is the reference image, and $P$ is the resulting pictorial chart that blends $R$ and $S$. We generated 10,000 examples of bar, line, and pie charts each to form the whole dataset.

\begin{figure}[t]
  \centering
  \includegraphics[width=\linewidth]{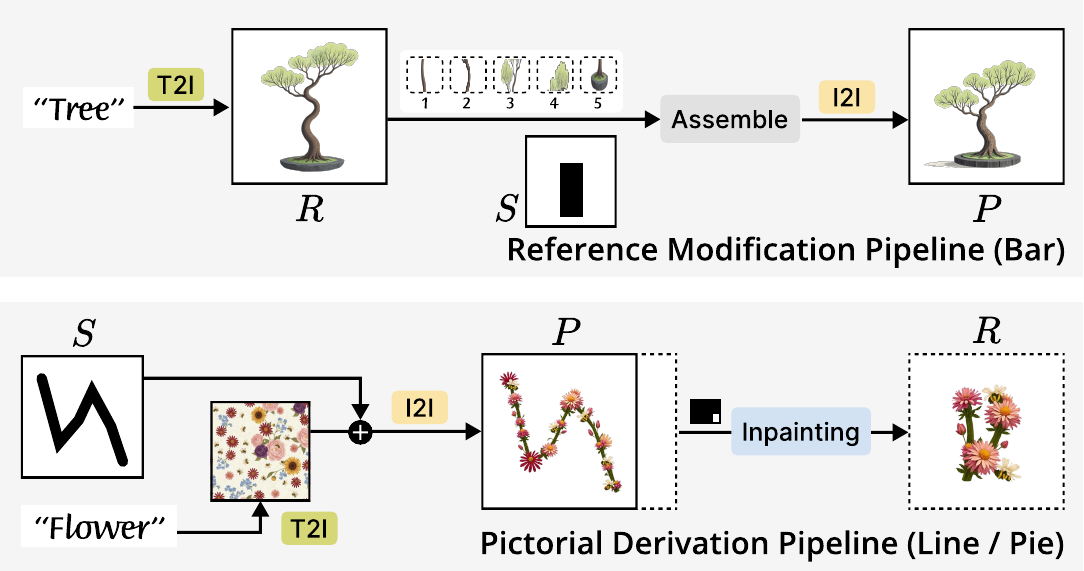} 
  \caption{Pipelines for \textsc{ChArtist-30K} Dataset Construction.}
  \label{fig:dataset}
\end{figure}

Different chart types require distinct generation pipelines, as the deformation needed to adapt the reference image $R$ with the geometric properties in $S$ varies substantially across charts.
For example, bar charts require the height of an object to be adjusted, while line and pie charts require more dramatic deformation to match the target topologies like curvature and angles.
Thus, we propose two specialized generation pipelines, which are illustrated in Fig.~\ref{fig:dataset}.
They generalize by covering two encoding rules: Pipeline (1) applies to charts with single linear or rectilinear encodings (e.g., bar, histogram, box, treemap, heatmap), while Pipeline (2) applies to charts with multi-dimensional or angular encodings (e.g., line, pie/donut, scatter, radar).

\subsection{Reference-Modification-based Pipeline}
For bar charts, this pipeline follows an $(R, S) \rightarrow P$ process: we first generate a reference image $R$, and then adjust it according to the height constraints specified by the chart skeleton $S$ to form the pictorial chart $P$.

\noindent\textbf{Reference Object Generation.}
We first generate the reference object $R$ using text-to-image (T2I) model with a prompt emphasizing a single object on a clean background. We then remove the background with BiRefNet~\cite{zheng2024birefnet} to obtain its precise height. Such a result is expected to be equivalent to a pictorial bar chart as well.

\noindent\textbf{Pictorial Chart Assembly.}
We vertically divide generated $R$ into $K=5$ equal-sized grids ${g_1, ..., g_K}$.
Our goal is to assemble these grids to align with the target height defined in $S$.
To identify structurally editable grids, we compute the Structural Similarity Index (SSIM) between all grid pairs $(g_i, g_j)$ where $i \neq j$.
Grids with higher mean SSIM represent repetitive textures that can be safely extended or trimmed without disrupting the object’s visual coherence.
We rank all grids by their mean SSIM scores to obtain an editability priority queue.
Guided by this ranking, we either replicate the highest-ranked grid or remove lower-ranked grids to match the height specified in $S$.
Finally, the assembled image is injected into an image-to-image (I2I) pipeline for refinement, producing the final seamless pictorial chart $P$.
This design is robust even for non-vertical objects.

\subsection{Pictorial-Derivation-based Pipeline}
Directly warping a canonical object $R$ to fit the curved skeleton $S$ is ill-posed (e.g., fitting a flower into a line chart). Thus, we adopt a reverse pipeline $S \rightarrow P \rightarrow R$: we first generate a plausible pictorial chart $P$ that matches the shape of $S$, and then derive the reference object $R$.

\noindent\textbf{Pictorial Chart Generation.} 
We begin by generating a textured background that contains repetitive patterns of a reference scene using a T2I model. The chart skeleton $S$ is then used as a binary mask to crop the chart-shaped region, forming an initial $P$.
This initial $P$ often contains incompleted objects in the chart region.
We refine it using an I2I model to form the final $P$ such that the objects obtain the missing details.

\noindent\textbf{Reference Object Derivation.}
To derive $R$, we employ a diptych prompting strategy~\cite{shin2025large} with an inpainting model: we append a blank panel to the right of $P$ and prompt the model to generate an object that matches the appearance of $P$ but with a natural, standalone shape. The filled region is cropped as the reference object $R$.
The result (i.e., the filled region) is cropped to serve as our reference object $R$. Noted that sometimes $R$ might still carries chart-like characteristics. To tackle this, we repeat the inpainting process using the latest $R$ as input to obtain a more natural object. Manual verification and filtering are applied at both stages to ensure visual consistency and data faithfulness.
\section{Method}
\label{sec: method}

Our method builds on a pretrained Diffusion Transformer (DiT) model to blend visual imagery with chart data.
Given an input chart, we support both text-driven and reference-image-driven generation to allow different levels of customization.
The output image should align with the structure defined by the chart while conforming to the visual semantics specified by the text or reference image.
To achieve this, we first introduce a skeleton-based representation that explicitly encodes the chart structure to guide spatial control.
We then train two LoRA modules to achieve spatial and subject control respectively.
However, we find that naively combining LoRAs in parallel leads to cross-condition interference, where conflicts between the two controls lead to degraded structure misalignment and noticeable style leakage; see Fig.~\ref{fig:challenge}.
To address this, we propose a cascaded conditioning inference paradigm that establishes an explicit dependency between conditions, dynamically gating the subject’s influence by the spatial signal.

\begin{figure}[t]
  \centering
  \includegraphics[width=\linewidth]{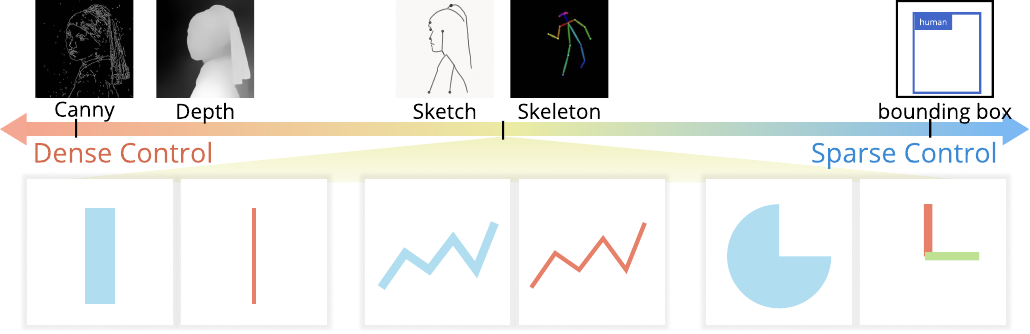} 
  \caption{The spectrum of control representation based on their complexity. Top: Existing control representations for natural image; Bottom: Our proposed skeleton-based representations for pictorial chart generation, where charts (blue) are represented as skeletons (red and green lines).}
  \label{fig:sk_method}
\end{figure}

\subsection{Data-Driven Control Representation}

To unify spatial and subject control, the control representation for pictorial chart generation must faithfully encode data while maintaining visual flexibility for stylistic adaptation.
Existing control representations often fail to meet this requirement, tending either to be overly dense or overly sparse.
As we conceptualize on a spectrum (Fig.~\ref{fig:sk_method}), dense pixel-level conditions (e.g., depth, Canny) are overly complex, leaving little room for stylistic infusion, whereas sparse conditions (e.g., bounding boxes) are too weak to guide internal structures.
We follow the concise design of sketch and skeleton, and propose a skeleton-based representation that occupies a sweet spot in the spectrum, where it 1) faithfully encodes the primary data-encoding dimension, and 2) is structurally minimal to enable flexible semantic or stylistic injection.

Specifically, we define three types of chart-type-specific skeletons: a single vertical line represents each bar in bar charts, a polyline traces the trend in line charts, and two colored radial lines indicate the clockwise start and end angles of each slice in pie charts.

\begin{figure*}[t]
  \centering
  \includegraphics[width=\linewidth]{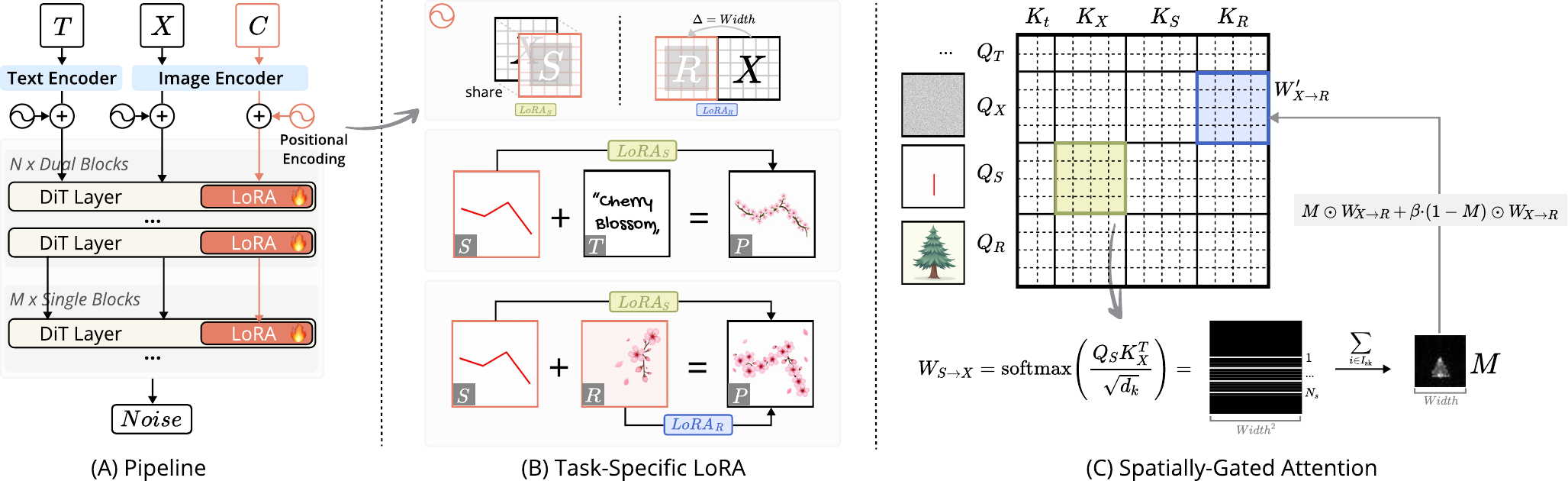} 
  \caption{The whole architecture of ChArtist consists of (A) a pretrained DiT-based diffusion model with two conditional-LoRAs with different positional encoding on the image inputs, where (B) one is for spatial control ($LoRA_S$) and another for subject control ($LoRA_R$). (C) To avoid the interference from $LoRA_R$ to $LoRA_S$, we employ a Spatially-Gated Attention mechanism at inference time.}
  \label{fig:pipe}
\end{figure*}

\subsection{Learning Spatial and Subject Control}
\paragraph{Training Pipeline.}
We adapt a DiT-based diffusion model following a conditional-LoRA architecture~\cite{tan2025ominicontrol, wang2025unicombine}. The image condition token $C$ (\textit{i.e.}, either chart skeleton $S$ for spatial or reference image $R$ for subject) is concatenated with the text tokens $T$ and noisy image tokens $X$ to form a unified input sequence $[T, X, C]$, as illustrated in Fig.~\ref{fig:pipe} (A). During forward process, $T$ and $X$ are handled by the frozen pretrained backbone, whereas $C$ is processed by trainable LoRA adapters. The unified sequence allows flexible interactions among textual, latent, and conditional tokens through multimodal attention.

\paragraph{Task-Specific LoRA.}
We train two task-specific LoRA adapters, $LoRA_S$ for spatial control and $LoRA_R$ for subject control. They can be used independently or jointly to support different sources of semantic information, such as concept semantics from text prompts or appearance features from reference images (Fig.~\ref{fig:pipe} B).
Their training differs in two aspects. First, $LoRA_S$ is trained on skeleton–pictorial pairs $(S, P)$, while $LoRA_R$ is trained on reference–pictorial pairs $(R, P)$.
Second, spatial control requires spatial alignment with the latent $X$, whereas subject control does not. To unify them within the same framework, we adopt a RoPE-based position-aware strategy~\cite{su2024roformer, tan2025ominicontrol}. The skeleton $S$ shares positional indices with the latent tokens $X$, while the reference $R$ is shifted by an offset $\Delta$ to the side of $X$ in latent space.

\subsection{Dual-Control Inference}
\label{ssec: dual_method}
\paragraph{Challenge of Merging Multiple LoRAs.}

\begin{figure}[t]
  \centering
  \includegraphics[width=\linewidth]{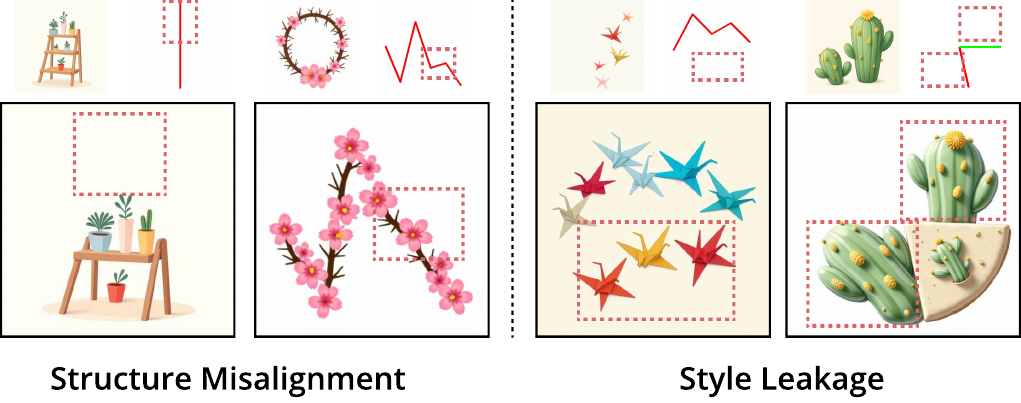} 
  \caption{Artifacts observed when merging multiple LoRAs in parallel. $LoRA_R$ can distort the spatial control and disobey the chart structure (structure misalignment), and can also introduce extra visual elements beyond the spatial constraints (style leakage).}
  \label{fig:challenge}
\end{figure}

Conventional multi-condition control typically composes LoRAs in parallel, treating each signal as an independent input. However, such parallel composition introduces severe cross-condition interference, which is especially harmful for pictorial charts where spatial and subject constraints must be tightly coordinated. This interference causes the generated chart to misrepresent the underlying data, manifesting in two primary failure modes, as shown in Fig.~\ref{fig:challenge}: the result may fail to adhere to the skeleton structure (\textit{structure misalignment}), or it may fill the chart region correctly but still leak subject content into the background, making the chart visually ambiguous (\textit{style leakage}).

\paragraph{Spatially-Gated Attention.}
To resolve this interference, we move from a parallel-competition paradigm to a sequential-conditioning paradigm at inference time.
Pictorial charts inherently require an explicit dependency between spatial and subject control: the subject must be coordinated by the spatial constraint.
To enforce this dependency, we propose a training-free inference mechanism called \textit{Spatially-Gated Attention}. The core idea is to derive a spatial mask from the spatial condition and use it to dynamically gate the influence of the subject signal.

We first construct a spatial mask $M$. Since the skeleton $S$ is a sparse and abstract representation that does not directly encode the concrete object shape, we compute an attention map between the skeleton queries $Q_S$ and latent keys $K_X$:
\[
W_{S \to X} = \textup{softmax}\!\left( \frac{Q_S K_X^T}{\sqrt{d_k}} \right),
\]
where each element $(W_{S \to X})_{i,j}$ represents the probability that latent token $j$ aligns with skeleton token $i$.
To identify the set of data-encoding skeleton tokens, denoting as $I_S \subseteq \{1,\dots,N_S\}$, we map foreground pixel coordinates (e.g., colored pixels) from the skeleton image to their corresponding 1D token indices in the latent sequence via coordinate downsampling and flattening. The spatial mask $M$ is then computed by aggregating the attention over this set $I_S$:
\[
M = \sum_{i \in I_S} (W_{S \to X})_i.
\]
This mask is then applied to gate the subject attention. The original subject attention weights $W_{X \to R}$ (from $Q_X$ to $K_R$) are replaced by gated weights $W'_{X \to R}$:
$$W^\prime_{X \to R} = M \odot W_{X \to R} + \beta \cdot (1 - M) \odot W_{X \to R},$$
where $\odot$ denotes element-wise multiplication, and $\beta$ controls the degree of subject expressiveness in background regions.
Finally, a normalization step is performed to restore the probability distribution.

\section{Experiments}
\subsection{Settings}
\noindent \textbf{Training.}
We follow the default settings of \textsc{OmniControl}~\cite{tan2025ominicontrol} and fine-tune \textsc{FLUX.1-dev} on our \textsc{ChArtist-30K} using LoRA.
For each chart type, we train two LoRA adapters for spatial control and subject control with rank $16$ at $512\times512$ resolution.
All experiments are conducted on $2\times$ NVIDIA A100 (80GB) GPUs.
Each model is trained for $25{,}000$ iterations per task.

\noindent \textbf{Evaluation.}
We evaluate the model on two tasks: spatially aligned control and subject-guided control. A method is considered robust if it can handle a wide range of visual concepts while adapting to different chart structures.
To assess this, we generate 500 evaluation images per task for each chart type, using prompts produced by ChatGPT that span 30 major categories (e.g., plants, animals, buildings, sports, etc.).
For the subject-guided task, the reference image is produced by \textsc{FLUX.1-schnell}.
$\beta$ is set to 0.6.
More Evaluation details can be found in Appendix~\ref{sec:training_detail}.

\noindent \textbf{Task 1: Spatially Aligned Only Task.}
We compare our method against baselines with various spatial control representations: ControlNet (Canny/Depth)~\cite{zhang2023adding}, SDEdit~\cite{meng2022sdedit}, and Inpainting~\cite{rombach2021highresolution}.
All baselines use FLUX.1 as the backbone.
For SDEdit relying on the colored strokes, we prompt ChatGPT to return a theme-related color based on the text prompt and apply it to the strokes.

\noindent \textbf{Task 2: Subject-Guided Task.}
We compare against the combination of ControlNet (Canny/Depth) and IP-Adapter~\cite{ye2023ip-adapter}, Paint-by-Example~\cite{yang2022paint}, and current advanced image editing models including Qwen-Image-Edit~\cite{wu2025qwenimagetechnicalreport},  Nano Banana~\cite{geminiteam2025geminifamilyhighlycapable}, and GPT-Image-1~\cite{openai_gptimage1}
. We only evaluate the similarity between the semantics of the references and generations \textit{within} the chart boundaries.

\subsection{Evaluation Metrics}
\begin{figure}[t]
  \centering
  \includegraphics[width=\linewidth]{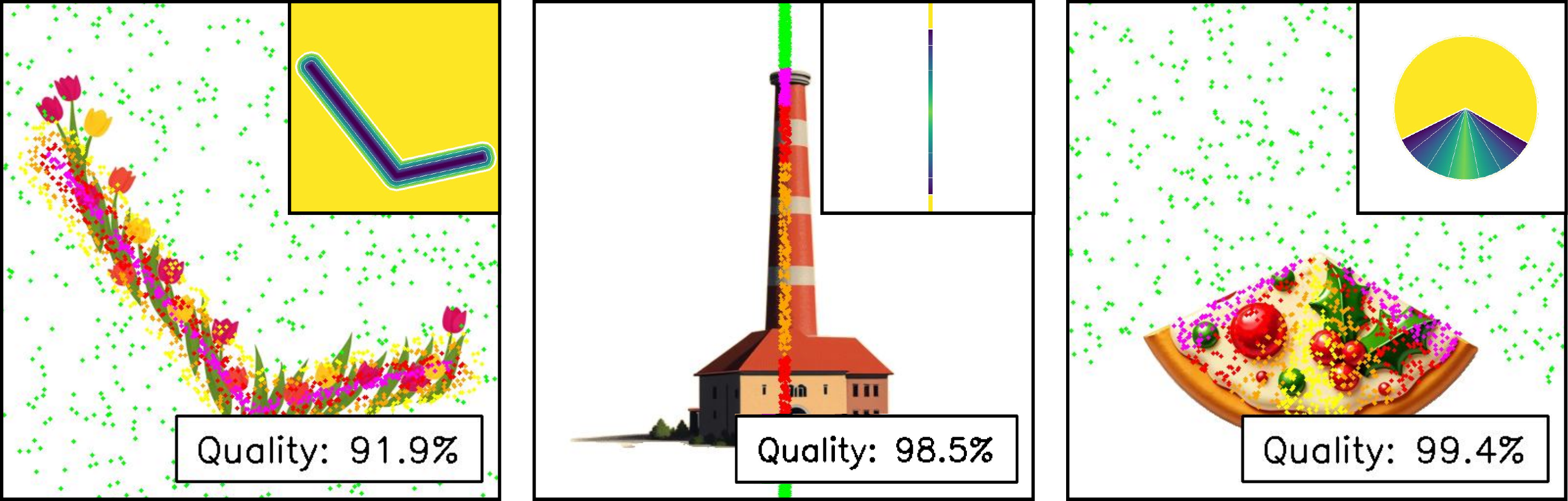} 
  \caption{Illustrations of the data accuracy metric. We construct a distance field based on the chart skeleton. Then we randomly sample the points based on the ranges on the distance field, and calculate the accuracy based on a weighted F1 Score.}
  \label{fig:data_acc_vis}
\end{figure}

\paragraph{Data Accuracy Metric.}
We design a \textit{structure-aware} \text{F1} Score to evaluate structural alignment between generated pictorial charts and their skeletons.
As chart skeletons are sparse, conventional spatial metrics such as IoU fail to capture structural alignment.  
To address this, we construct a distance field along the chart’s data-encoding dimension, where larger distances indicate a stronger geometric bias (see line chart example in Fig.~\ref{fig:data_acc_vis}).
We divide the chart area into multiple regions defined by distance and sample points within each region to approximate precision and recall.  
Each region is assigned a weight based on its spatial importance, forming a weighted score that reflects both geometric accuracy and structural completeness.
The implementation details can be found in \ref{supp:data_accuracy}.
Importantly, our weighting scheme is guided by the data-encoding type:
\begin{itemize}
    \item \textbf{Line chart:}
    Regions closer to the data trajectory receives higher weights to emphasize positional accuracy.
    \item \textbf{Bar chart:}  
    weights peak near the bar endpoints that represent height information.
    \item \textbf{Pie chart:}
    we assign higher weights to the radial division lines that determine the angle of the pie.
\end{itemize}

\paragraph{Semantic Alignment and Image Quality.}
Text alignment is measured using CLIP Text score.
For evaluating visual consistency between reference and generated image, we report CLIP-Image and DINO~\cite{caron2021emerging}.
To reduce background bias, visual consistency scores are computed on the cropped data-encoding region rather than on the full image.
Image quality is measured using CLIP-IQA~\cite{wang2022exploring}, MAN-IQA~\cite{yang2022maniqa}, and MUSIQ~\cite{ke2021musiq}.

\paragraph{User Perceptual Study.}
To assess human preferences regarding data faithfulness and semantic quality, we conducted an online comparison study with 300 participants.
Participants were evenly assigned to bar, line, or pie question sets, each containing 50 data-accuracy ranking questions and 50 semantic-alignment ranking questions.
Each question asked participants to rank four methods, with the method order randomized to reduce bias; in total, we obtained 100 responses for each of 300 unique questions.
Each participant was compensated \$15.

\subsection{Results}

\begin{figure}[t]
  \centering
  \includegraphics[width=0.99\linewidth]{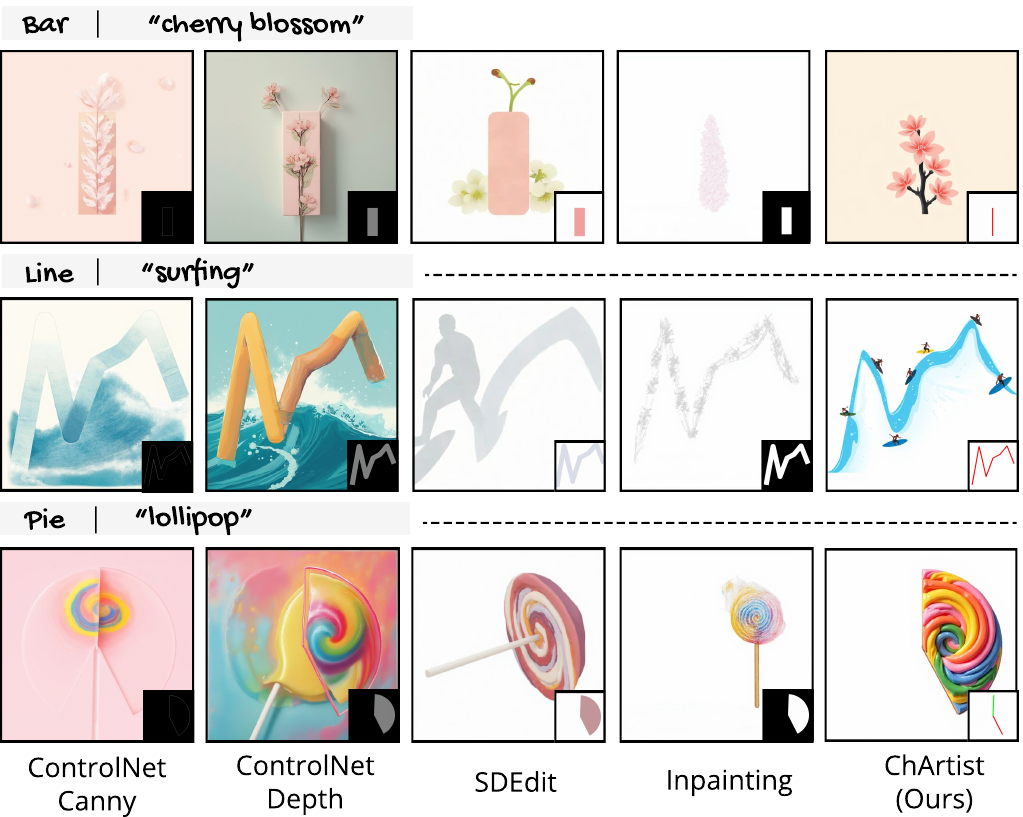} 
  \caption{Result of spatially aligned evaluation with different control representations. (Task 1)}
  \label{fig:eva_sk}
\end{figure}

\paragraph{Spatially Aligned Evaluation with different Control Representations.}
We show the qualitative results in Fig.~\ref{fig:eva_sk} and quantitative results in Tab.~\ref{tab:chart_compare_spatial_transposed}, with more results can be found in ~\ref{supp:more_result}.
As shown in the figure, baseline methods exhibit a clear trade-off between preserving chart structure and aligning with textual semantics.
For instance, ControlNet rigidly follows its Canny or depth conditions, often pushing semantic content into the background instead of the chart regions, compromising data fidelity.
Similarly, SDEdit, despite having good color initialization, adapts the global appearance to match the prompt but frequently distorts underlying data information.
Inpainting performs well on simple chart types such as bar charts, but struggles when deformation is required, often failing to fully populate the chart region, as illustrated by the incomplete lollipop-shaped pie chart.
Furthermore, because inpainting is operated on a blank background without visual context, its results tend to be less expressive and often degrade into sketch-like appearances.
In contrast, ChArtist achieves a far more harmonious balance, integrating semantic visual elements into chart structures without sacrificing expressiveness.
It does not rely on rigid contour adherence as ControlNet does, preventing content from being confined to hard structural boundaries.
Meanwhile, its spatial guidance is substantially stronger than weak image-conditioned approaches like SDEdit or Inpainting, which often drift away from the intended chart shape or become unrecognizable.

This observation is further supported by the quantitative results in Tab.~\ref{tab:chart_compare_spatial_transposed}.
ChArtist consistently delivers stable overall performance across chart types by evaluating both data accuracy and text alignment.
Notably, it outperforms other methods on CLIP-T on both types of chart, demonstrating it can incorporate the visuals while respecting structural constraints.
Note that while some baselines excel in isolated cases (e.g., Inpainting achieves highest bar chart data accuracy), they do so at the cost of yielding a low CLIP-T score.

\begin{table}[tb]
  \centering
  \caption{Quantitative comparison of spatially aligned evaluation.}
  \label{tab:chart_compare_spatial_transposed}

  \begin{adjustbox}{max width=\columnwidth}
  {\small
  \begin{tabular}{@{}cc|ccccc@{}}
    \toprule
    \multicolumn{2}{c|}{} & \multicolumn{5}{c}{\textbf{Method}} \\
    \cmidrule(lr){3-7}
    \multicolumn{2}{c|}{} & ControlNet-C & ControlNet-D & SDEdit & InPainting & ChArtist (Ours) \\
    \midrule
    \multirow{3}{*}{\textbf{Bar}} 
      & Data Acc $\uparrow$   & 0.741 & \worst{0.686} & 0.774 & \best{0.923} & 0.894 \\
      & CLIP-T $\uparrow$ & 0.249 & 0.243 & 0.233 & 0.231 & \best{0.304} \\
      & Avg $\uparrow$    & 0.495 & \worstavg{0.465} & 0.504 & 0.577 & \bestavg{0.599} \\
    \midrule
    \multirow{3}{*}{\textbf{Line}} 
      & Data Acc $\uparrow$   & 0.819 & 0.858 & 0.792 & \worst{0.754} & \best{0.920} \\
      & CLIP-T $\uparrow$ & 0.227 & 0.243 & 0.190 & \worst{0.179} & \best{0.247} \\
      & Avg $\uparrow$    & 0.523 & 0.550 & 0.491 & \worstavg{0.467} & \bestavg{0.584} \\
    \midrule
    \multirow{3}{*}{\textbf{Pie}} 
      & Data Acc $\uparrow$   & 0.725 & \worst{0.626} & \best{0.836} & 0.794 & 0.778 \\
      & CLIP-T $\uparrow$ & 0.136 & 0.158 & 0.190 & 0.217 & \best{0.252} \\
      & Avg $\uparrow$    & 0.430 & \worstavg{0.392} & 0.513 & 0.506 & \bestavg{0.515} \\
    \bottomrule
  \end{tabular}
  }
  \end{adjustbox}
\end{table}

\begin{table}[h]
  \centering
  \caption{Human evaluation results from controlled online study (300 participants). Average rank ranges from 1 (best) to N (worst).}
  \label{tab:controlled_online_study1}
  \resizebox{\columnwidth}{!}{
  \begin{tabular}{c c | c c}
    \toprule
    \textbf{Method} &
    \textbf{Spatial } &
    \textbf{Subject } &
    \textbf{Method} \\
    \midrule

    ControlNet-Canny 
      & 3.008
      & 2.914 \textcolor{bronzecolor}{(3rd)}
      & ControlNet-Canny  \\

    ControlNet-Depth 
      & 2.969 \textcolor{bronzecolor}{(3rd)}
      & 2.822 \textcolor{goldcolor}{(1st)}
      &  ControlNet-Depth  \\

    SDEdit
      & 2.897 \textcolor{goldcolor}{(1st)}
      & 3.085
      & Paint-by-Example \\

    Inpainting
      & 3.168
      & -
      & - \\

    ChArtist (Ours)
      & 2.957 \textcolor{silvercolor}{(2nd)}
      & 2.845 \textcolor{silvercolor}{(2nd)}
      & ChArtist (Ours) \\

    \bottomrule
  \end{tabular}}
\end{table}

\paragraph{Human Preference from Controlled Online Study.}
From Tab.~\ref{tab:controlled_online_study1}, our method demonstrates consistent performance (2nd place for both skeleton and reference questions), which corroborates our qualitative results that it offers the most balanced performance across both data accuracy and semantic alignment. 
More analysis and significance tests can be found in Appendix \ref{sec:sig_test}.

\begin{table*}[h]
  \centering
  \caption{Quantitative comparison result of dual control of spatial and subject (Task 1 + Task 2).}
  \label{tab:chart_compare_sub}
  \resizebox{\textwidth}{!}{
  \begin{tabular}{cccc|ccc|ccc|ccc}
    \toprule
    \multirow{2}{*}{\textbf{Method}} &
    \multicolumn{3}{c|}{\textbf{Bar}} &
    \multicolumn{3}{c|}{\textbf{Line}} &
    \multicolumn{3}{c|}{\textbf{Pie}} &
    \multicolumn{3}{c}{\textbf{Image Quality}} \\
    & \textbf{Data Acc} $\uparrow$ & \textbf{DINO} $\uparrow$ & \textbf{CLIP-I} $\uparrow$
    & \textbf{Data Acc} $\uparrow$ & \textbf{DINO} $\uparrow$ & \textbf{CLIP-I} $\uparrow$
    & \textbf{Data Acc} $\uparrow$ & \textbf{DINO} $\uparrow$ & \textbf{CLIP-I} $\uparrow$
    & \textbf{CLIP-IQA} $\uparrow$ & \textbf{MUSIQ} $\uparrow$ & \textbf{MAN-IQA} $\uparrow$ \\
    \midrule

    ControlNet-Canny + IP-Adater
      & 0.634 & 0.652 & 0.824
      & 0.728 & 0.613 & 0.758
      & 0.652 & 0.651 & 0.701
      & 0.674 & 67.38 & 0.487 \\

    ControlNet-Depth + IP-Adater
      & \worst{0.593} & 0.635 & 0.805
      & 0.683 & 0.615 & 0.752
      & 0.579 & 0.668 & 0.723
      & \worst{0.620} & \worst{54.66} & \worst{0.372}\\

    Paint-by-Example
      & 0.912 & \worst{0.586} & \worst{0.708}
      & \worst{0.513} & \worst{0.429} & \worst{0.615}
      & \worst{0.420} & \worst{0.495} & 0.634
      & 0.683 & 65.37 & 0.574 \\
\midrule
    Qwen-Image-Edit
      & 0.733 & 0.697 & 0.856
      & 0.574 & 0.621 & 0.732
      & \best{0.765} & 0.578 & \worst{0.621}
      & 0.660 & 63.18 & 0.567 \\

    Nano Banana
      & 0.727 & 0.731 & 0.812
      & 0.716 & 0.606 & 0.685
      & 0.546 & \best{0.692} & 0.727
      & 0.679 & 65.32 & 0.565 \\

    GPT-Img-1
      & 0.758 & 0.745 & 0.830
      & 0.628 & 0.679 & 0.756
      & 0.422 & 0.657 & 0.718
      & \best{0.712} & 67.98 & 0.581 \\

    \midrule
    \textbf{ChArtist (Ours)}
      & \best{0.931} & \best{0.837} & \best{0.892}
      & \best{0.905} & \best{0.728} & \best{0.815}
      & 0.753 & 0.689 & \best{0.747}
      & 0.657 & \best{69.35} & \best{0.589} \\

    \bottomrule
  \end{tabular}}
\end{table*}

\begin{figure*}[tb]
  \centering

  \newcommand{\leftw}{0.64\textwidth}  
  \newcommand{\rightw}{0.32\textwidth}

  \begin{subfigure}[t]{\leftw}
    \centering
    \includegraphics[width=\linewidth]{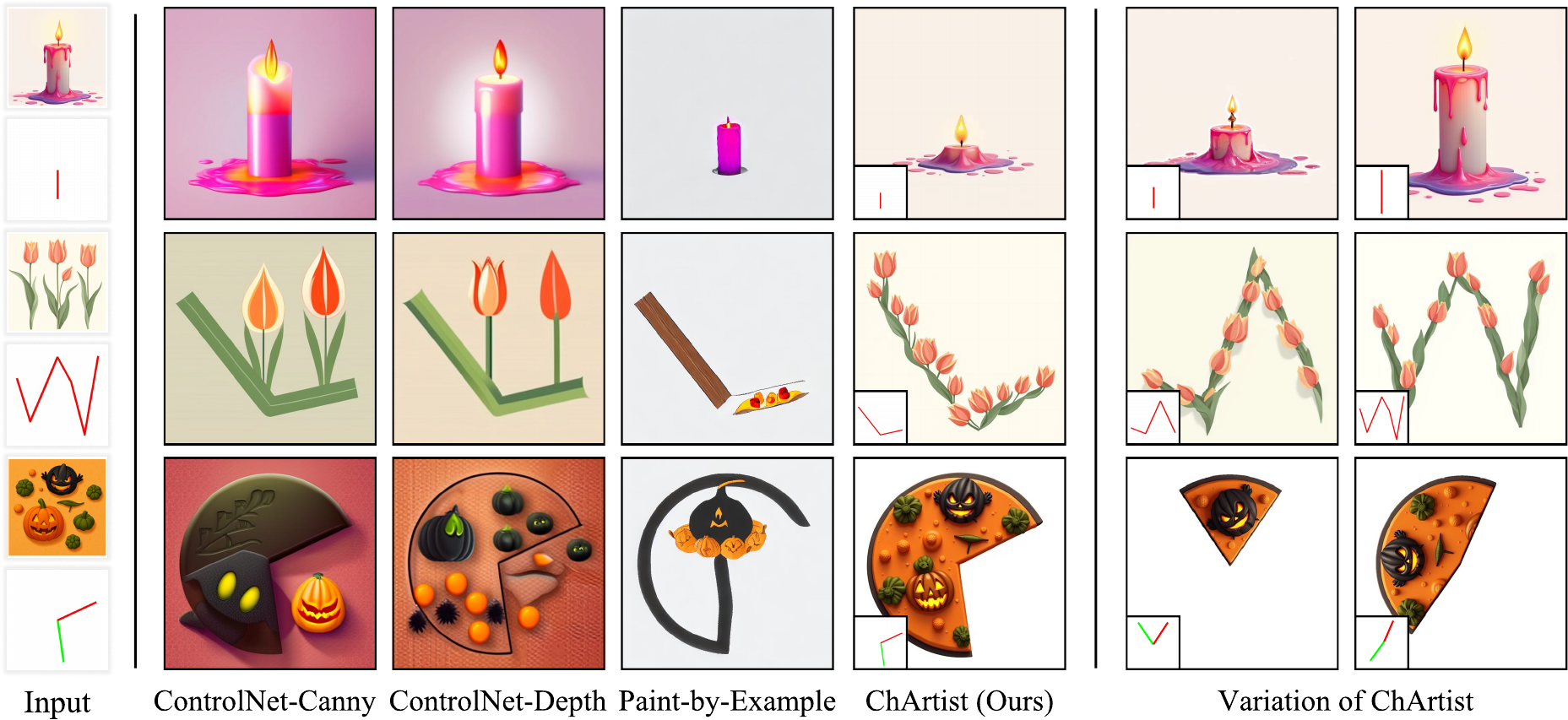}
    \caption{}
    \label{fig:eva_sub}
  \end{subfigure}
  \hfill
  \begin{subfigure}[t]{\rightw}
    \centering
    \includegraphics[width=\linewidth]{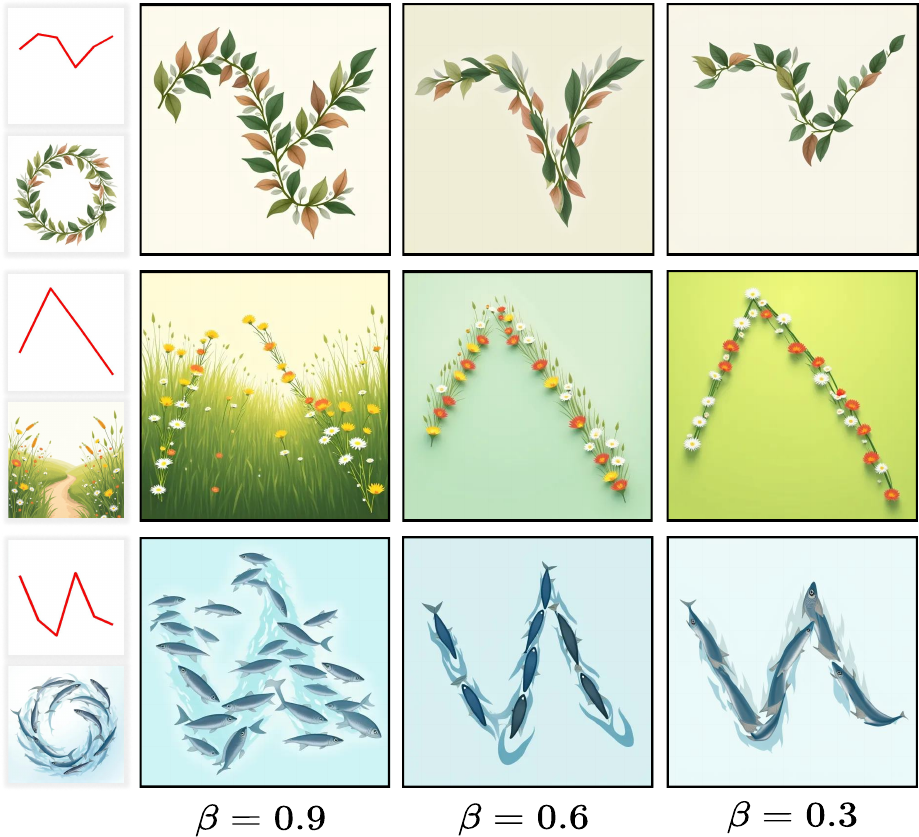}
    \caption{}
    \label{fig:eva_beta}
  \end{subfigure}

  \caption{(a) Dual-control generation results conditioned on both spatial structure and subject reference. (b) Results of ChArtsit with different $\beta$ showcases the importance of Spatially Gated Attention of dual control in ensuring the data accuracy.}
  \label{fig:eva_both}
\end{figure*}

\begin{figure}[t]
  \centering
  \includegraphics[width=0.99\linewidth]{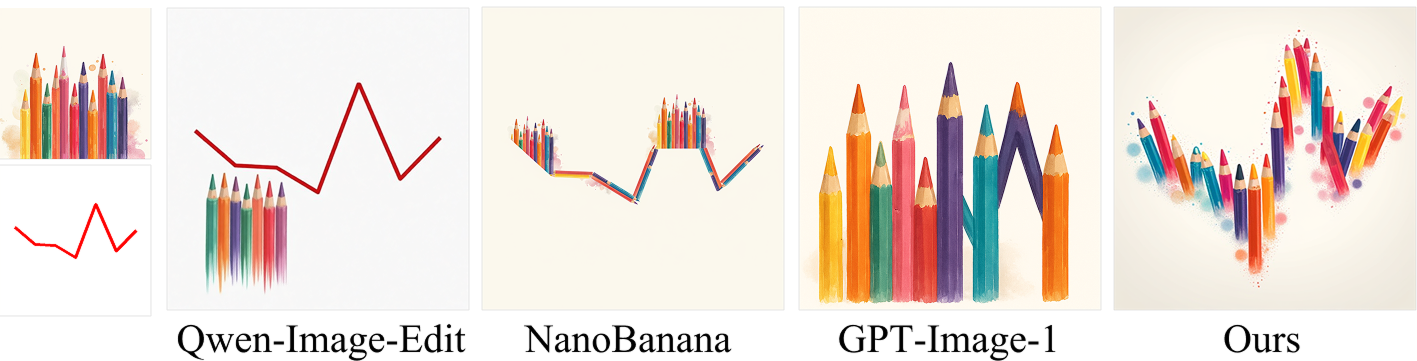} 
  \caption{Comparison with current SoTA image editing models.}
  \label{fig:result with nanobanana}
\end{figure}

\paragraph{Dual Control Evaluation.}
We present a quantitative comparisons in Tab.~\ref{tab:chart_compare_sub} and the qualitative analysis in Fig.~\ref{fig:eva_sub}, Fig.~\ref{fig:result with nanobanana}.
As shown in Tab.~\ref{tab:chart_compare_sub}, with more results can be found in ~\ref{supp:more_result}.
Our method consistently outperforms all baselines across all chart types, in terms of data accuracy, visual consistency with reference image, and image quality.
This observation of high data accuracy is the most evident in line charts, where ChArtist achieves a data accuracy of 0.905, surpassing the second-best method (0.728) significantly.
For visual consistency with the reference image (DINO and CLIP-I), ChArtist again achieves the best performance across all chart types.
Regarding overall image quality, we use the average score across three chart types; ChArtist still delivers high image quality, except for the CLIP-IQA score.
It is worth noting that although Paint-by-Example achieves a high data accuracy (0.912) for bar charts, this comes at the cost of visual consistency. Its DINO and CLIP-I scores are relatively low, indicating a failure to preserve the reference input's visual style.
The qualitative results in Fig.~\ref{fig:eva_sub} corroborate these quantitative findings. 
Baseline methods, despite their strong performances on natural images, generally struggle to blend the reference appearance into the correct chart regions. 
The bottom of Fig.~\ref{fig:eva_sub} and Fig.~\ref{fig:Teaser} presents variations of ChArtist generated from the same reference image under different structural inputs. The consistent preservation of appearance demonstrates that our dual-control framework enables both strong visual consistency and robust adaptability to structural variations.
The results in Fig.~\ref{fig:result with nanobanana} further demonstrate the potential of image editing models in preserving visual consistency. However, they still exhibit limited ability to faithfully follow structural constraints.

\begin{table}[t]
  \centering
  \caption{Ablation on control factor $\beta$ on line pictorial chart.}
  \label{tab:eva_beta}
  \resizebox{0.7\linewidth}{!}{%
  \begin{tabular}{c|c|ccc}
    \toprule
     $\beta$ & Data Acc $\uparrow$ & DINO $\uparrow$ & CLIP-T $\uparrow$ \\
    \midrule
     0.3 & \textbf{0.927} & 0.732 & 0.324 \\
     0.6 & 0.876 & 0.748 & \textbf{0.349} \\
     0.9 & 0.729 & \textbf{0.775} & 0.337 \\
    \bottomrule
  \end{tabular}%
  }
\end{table}
\paragraph{Subject Control Factor $\beta$ Ablation.}
We conduct an ablation study to evaluate the effectiveness of our spatially-gated attention mechanism (Sect.~\ref{ssec: dual_method}).
Fig.~\ref{fig:eva_both}(b) presents qualitative results as the suppression factor $\beta$ varies across different settings.
As $\beta$ decreases, the gating more strongly suppresses attention to non-chart regions.
This helps prevent reference-image content from leaking into the background;
for example, with $\beta=0.9$, the model tends to copy the the background of reference image, whereas lowering $\beta$ effectively mitigates this issue (second row of Fig.~\ref{fig:eva_both} (b)).
Meanwhile, stronger suppression encourages the model to focus more on the designated chart region, improving data accuracy, as shown by the result with $\beta$ set to $0.3$.
The quantitative results in Tab.~\ref{tab:eva_beta} corroborate these observations, showing how different $\beta$ values affect data accuracy.
Besides, they also show a trade-off between data accuracy and visual consistency, where the best data accuracy is not accompanied by the highest visual consistency score.

\section{Conclusion}

We introduce ChArtist, a pipeline for generating pictorial charts with text and image references.
We address the core challenge of balancing data faithfulness and visual expressiveness by providing both spatial and subject-level control, which can be used independently or composed jointly.
We propose a skeleton-based representation, a minimal abstraction that encodes only the core data dimensions, preserving freedom for visual integration.
We train spatial and subject LoRA separately and employ Spatially-Gated Attention during inference to mitigate their mutual interferences.
We support our method with a large-scale dataset, ChArtist-30K, and a data accuracy metric for quantifying the faithfulness of generated charts.
We demonstrate its usability and compatibility in the current design workflow, see ~\ref{supp:app}.
We hope that it will unlock the door for further research of image generation to empower data-driven visual storytelling.



{
    \small
    \bibliographystyle{ieeenat_fullname}
    \bibliography{main}
}

\clearpage
\clearpage
\setcounter{page}{1}
\maketitlesupplementary
\setcounter{section}{0}
\setcounter{figure}{0}
\setcounter{table}{0}
\setcounter{equation}{0}

\renewcommand{\thesection}{\Alph{section}}
\renewcommand{\thefigure}{S\arabic{figure}}
\renewcommand{\thetable}{S\arabic{table}}
\renewcommand{\theequation}{S\arabic{equation}}

\section{Evaluation}
\label{sec:supp-eva}

\subsection{Data Accuracy Evaluation Details}
\label{supp:data_accuracy}
We aim to measure how faithfully the generated image represents data using a structure-aware F1 score.

\noindent\textbf{Preprocessing.}
Before scoring, we first remove the background of each pictorial chart and apply a slight blur to its alpha channel to handle stylistic variance, such as hollow or soft strokes.  
For most chart types, we then collect all pixels with $\alpha > 0$ as the foreground region of the generated chart.  
For bar charts, however, we use the bounding box of the non-transparent region as the effective area,  
since slanted objects (e.g., Pisa tower) may not fully overlap with the skeleton despite being visually aligned.

\noindent \textbf{Sampling-based F1 score.} We randomly sample points from both the skeleton and the generated chart to approximate \textbf{precision} and \textbf{recall}:  
\begin{equation}
\text{Precision} = \frac{|P \cap S|}{|P|}, \quad
\text{Recall} = \frac{|P \cap S|}{|S|}.
\end{equation}
\noindent Here, $P$ and $S$ denote the sampled point sets from the generated chart and the skeleton, respectively.  
The overall score is computed as the harmonic mean of both terms:  
\begin{equation}
F1 score = 
\frac{2 \times (\text{Precision} \times \text{Recall})}
{(\text{Precision} + \text{Recall}) + \epsilon}.
\end{equation}

\noindent \textbf{Weighted Region Scheme.}  
Since points proximity to the skeleton indicates varying importance,  
we introduce multi-level weighted regions based on distance.  
for instance, for each region $i$, we sample a fixed number of points and compute weighted precision as:  
\begin{equation}
\text{Precision}_{\text{w}} = 
\frac{\sum_i w_i \times |P_i \cap S|}
{\sum_i w_i \times |P_i|}, 
\end{equation}

\noindent \textbf{Structure-aware Adaptation.}  
We further adapt the weighting scheme to each chart type guided by their data encoding, ensuring that regions most relevant to data semantics receive higher importance.  

\begin{figure}[h]
\centering
\begin{subfigure}{0.48\linewidth}
\centering
\includegraphics[width=\linewidth]{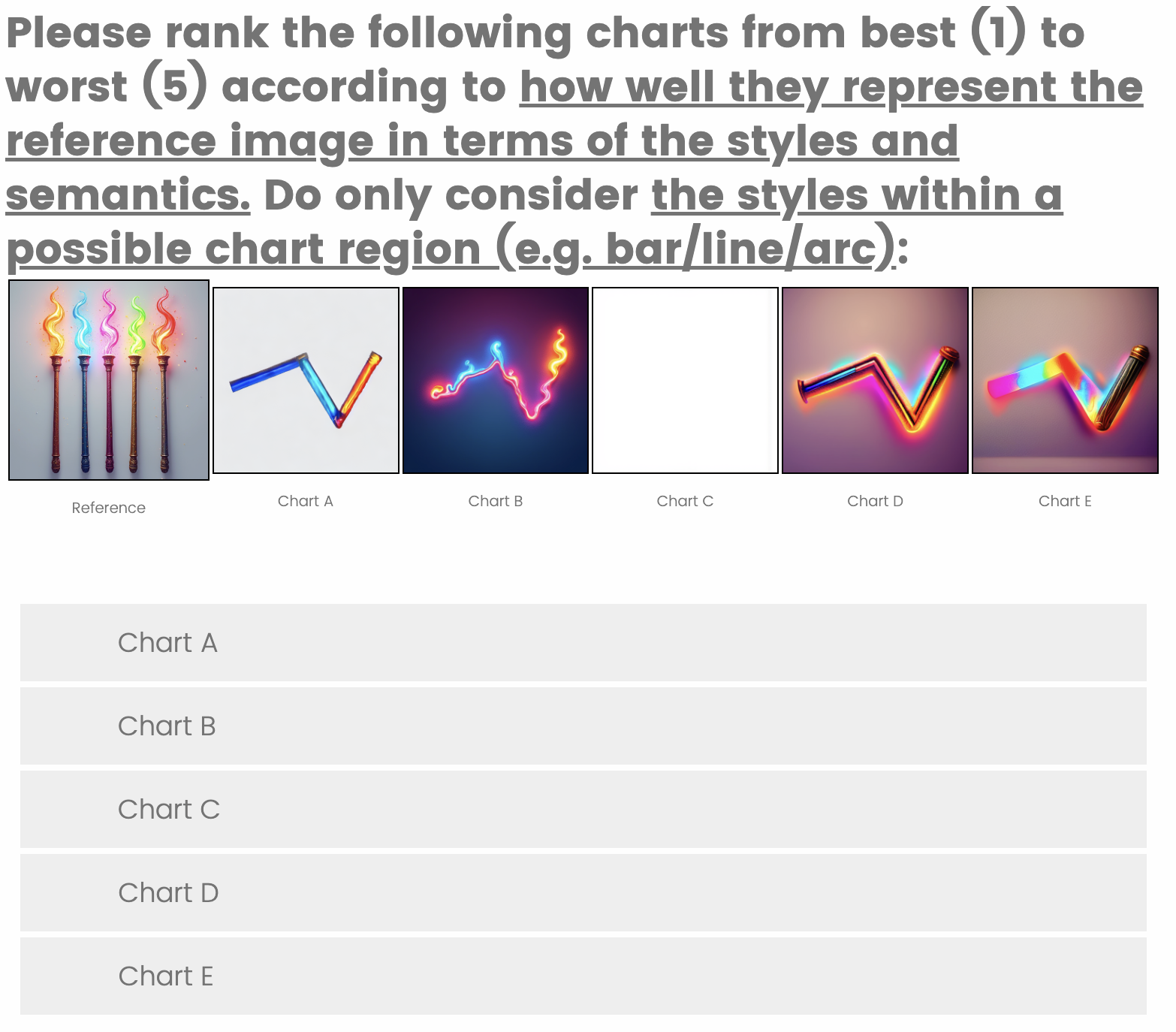}
\caption{Reference question sample}
\label{fig:supp-reference-question}
\end{subfigure}
\hfill
\begin{subfigure}{0.48\linewidth}
\centering
\includegraphics[width=\linewidth]{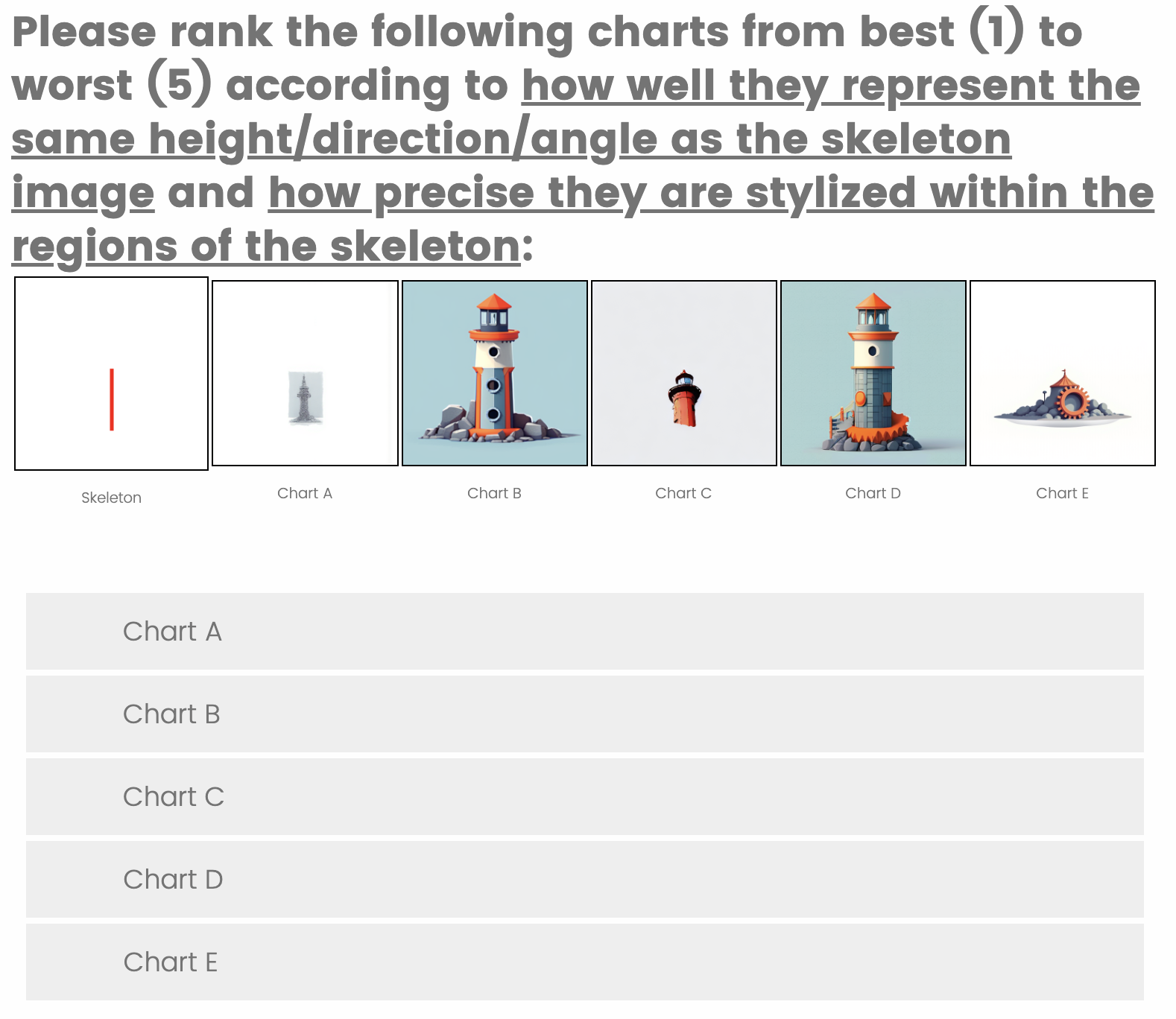}
\caption{Skeleton question sample}
\label{fig:supp-skeleton-question}
\end{subfigure}
\caption{Controlled online study questions for evaluating (a) style and semantic representation quality, and (b) skeleton alignment accuracy and precision.}
\label{fig:supp-user-study-questions}
\end{figure}

\subsection{Significance Tests}
\label{sec:sig_test}

For the controlled online study results, as shown in Fig.~\ref{tab:controlled_online_study}, a Friedman test revealed no significant difference among methods for Skeleton-condition questions $(\chi^2(4) = 4.02, p = 0.403)$, indicating comparable perceived quality compared against the structure of the chart skeleton. For Reference-condition questions, where participants had target reference image, the difference was significant $(\chi^2(4) = 24.52, p < 0.001)$. Post-hoc Wilcoxon signed-rank tests with Bonferroni correction $(\alpha = 0.005)$ identified significant pairwise differences. Specifically, Chartist significantly outperformed InContext $(p < 0.001)$, on par with the other top-ranked methods (Depth and Canny), and was among the top two methods in average ranking $(avg = 2.85)$.

While structural control (Skeleton questions) produced minimal perceptual differentiation, Chartist maintained strong relative preference, demonstrated by their average ranks, under conditions requiring \textit{both} faithful spatial correspondence and visual detail—suggesting that its control representation satisfies a niche that prefers a balance of semantic alignment and pictorial fidelity.

\begin{table}[tb]
  \centering
  \caption{Human evaluation results from controlled online study (300 participants). Average rank ranges from 1 (best) to N (worst). 
  }
  \label{tab:controlled_online_study}
  \resizebox{\columnwidth}{!}{
  \begin{tabular}{lccccc}
    \toprule
    \multirow{2}{*}{\textbf{Method}} &
    \multicolumn{2}{c}{\textbf{Spatial (150 Questions)}} &
    \multicolumn{2}{c}{\textbf{Subject (150 Questions)}} &
    \multirow{2}{*}{\textbf{Method}} \\
    \cmidrule(lr){2-3}\cmidrule(lr){4-5}
    & \textbf{Avg Rank} & \textbf{Kendall $\tau$} &
    \textbf{Avg Rank} & \textbf{Kendall $\tau$} &
    \\
    \midrule
    \textbf{ChArtist (Ours)} & 2.957 \textcolor{silvercolor}{(2nd)} & 0.052 & 2.845 \textcolor{silvercolor}{(2nd)} & 0.080 & ChArtist (Ours) \\
    SDEdit & 2.897 \textcolor{goldcolor}{(1st)} & 0.070 & 3.085 & 0.086 & Paint-by-Example \\
    ControlNet-Depth & 2.969 \textcolor{bronzecolor}{(3rd)} & 0.058 & 2.822 \textcolor{goldcolor}{(1st)} & 0.057 & ControlNet-Depth \\
    ControlNet-Canny & 3.008 & 0.051 & 2.914 \textcolor{bronzecolor}{(3rd)} & 0.069 & ControlNet-Canny \\
    Inpainting & 3.168 & 0.074 & -- & -- & -- \\
    \bottomrule
  \end{tabular}}
  \vspace{-1em}
\end{table}

\begin{figure*}[h]
  \centering
  \includegraphics[width=\linewidth]{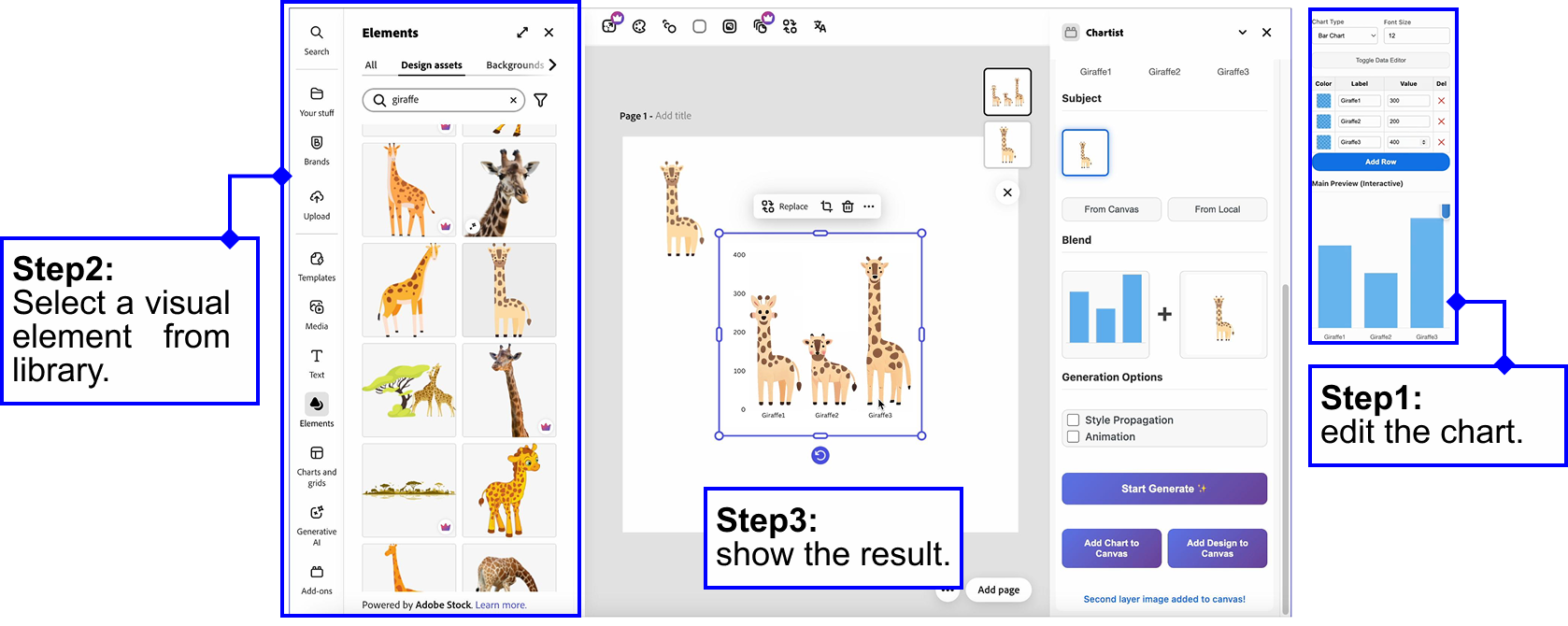} 
  \caption{Interface of ChArtist build in existing design platform. The left side is the visual library, the right side is chart and blending configuration. Canvas is in the central for user manipulation.}
  \label{fig:interface}
\end{figure*}

\subsection{Evaluation Details}
\label{sec:training_detail}

\paragraph{Implementation of Baseline.}
For the Spatially Aligned Only task (Task 1), all baseline methods use FLUX as the backbone model. For the subject-guided task, we use SDXL combined with ControlNet and IP-Adapter. Since the default line width is too thin to effectively present visual elements, we increase it to 30 pixels when creating the mask condition. Several baseline methods are sensitive to hyperparameters: we set the strength factor of SDEdit to 0.75 and the conditioning factor of ControlNet to 0.9.

\paragraph{Training Details}
For training both $LoRA_{R}$ and $LoRA_{S}$, we use the default prompt: ``this item, in a white background.''
The training dataset is highly diversified through the use of various style LoRAs, including cartoon, 3D, illustration, photorealistic, and more.

\subsection{Real-World Application}
\label{supp:app}
To illustrate the practical design value of ChArtist, we integrate it into a design software environment as a plugin (Fig.~\ref{fig:interface} and video in the supplementary). We observe that in most design platforms such as Adobe Express and Canva, visual design and chart creation are handled as two separate workflows. The visual-design workflow is rich, flexible, and supported by extensive asset libraries and recommendation systems, while chart creation is relatively limited, typically allowing customization only through simple color adjustments.
Built on Adobe Express, our goal is to demonstrate ChArtist’s ability to unify visual design and chart creation into a single workflow.
First, user can edit chart data in the right panel, and then select a visual element from the asset library. The blended result will be shown on the central canvas, which can be further added with axis and annotation.
We additionally incorporate external models to support a more complete design pipeline. For example, we employ StyleAligned to maintain stylistic consistency across visual elements, and we use an external image-to-video API to animate the final outputs.

\subsection{More Experimental Results}
\label{supp:more_result}
\begin{figure*}[t]
  \centering
  \includegraphics[width=0.78\linewidth]{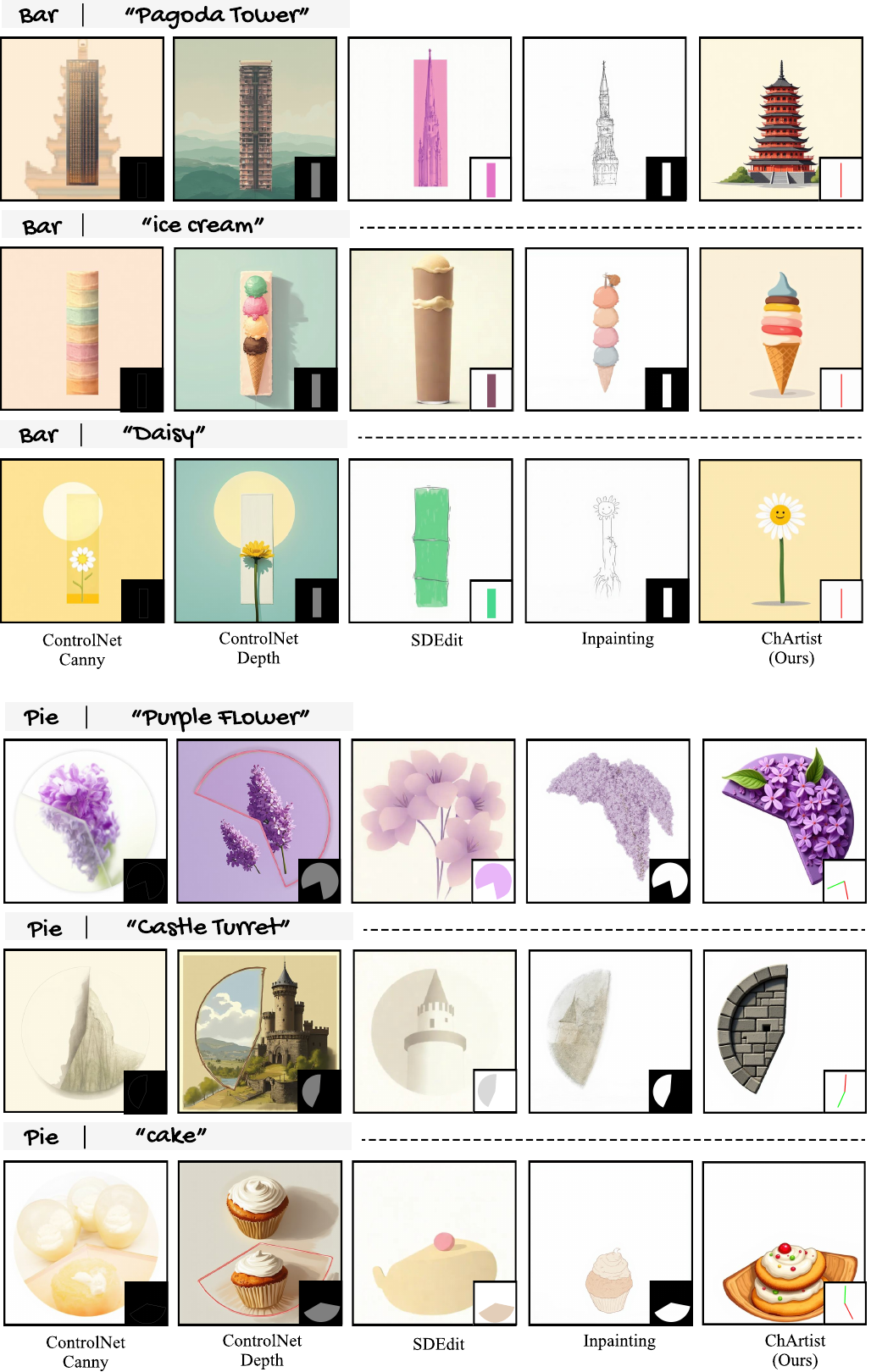} 
  \caption{Result of spatially aligned evaluation with different control representations. (Task 1).}
  \label{fig:supp_sk1}
\end{figure*}

\begin{figure*}[t]
  \centering
  \includegraphics[width=0.78\linewidth]{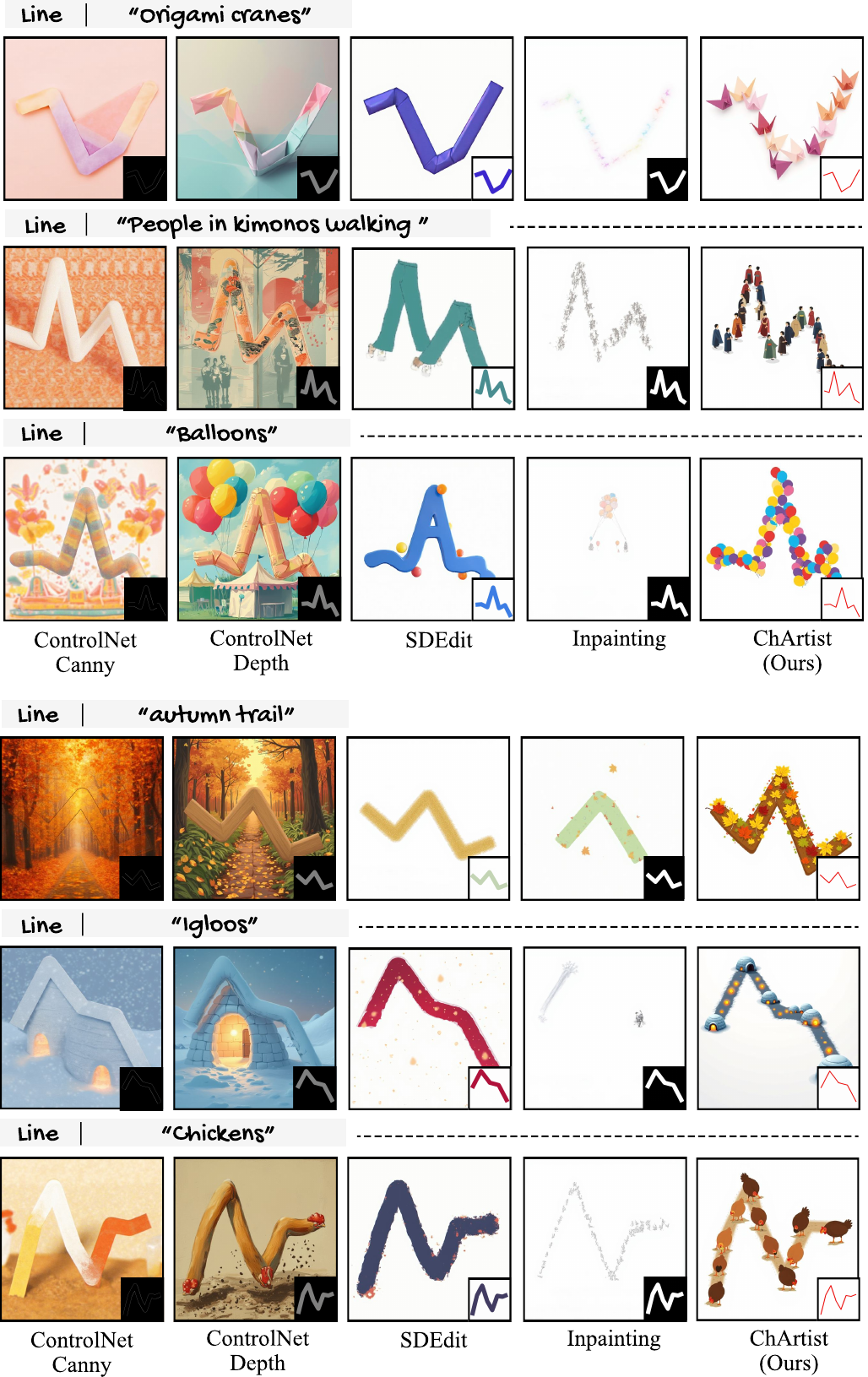} 
  \caption{Result of spatially aligned evaluation with different control representations. (Task 1).}
  \label{fig:supp_sk2}
\end{figure*}

\begin{figure*}[t]
  \centering
  \includegraphics[width=0.78\linewidth]{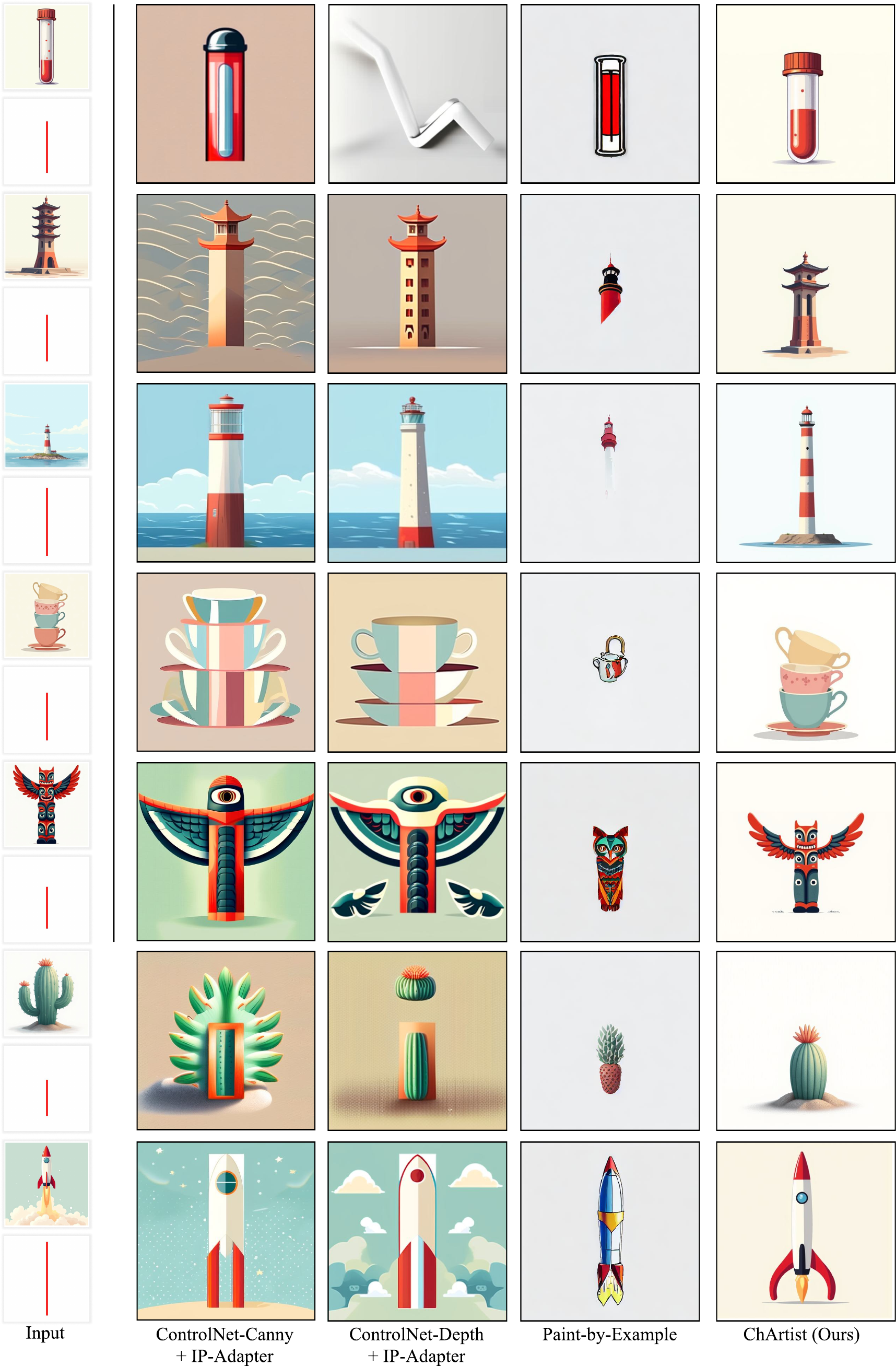} 
  \caption{Dual-control generation bar pictorial results conditioned on both spatial structure and subject reference.}
  \label{fig:supp_sub_bar}
\end{figure*}

\begin{figure*}[t]
  \centering
  \includegraphics[width=0.78\linewidth]{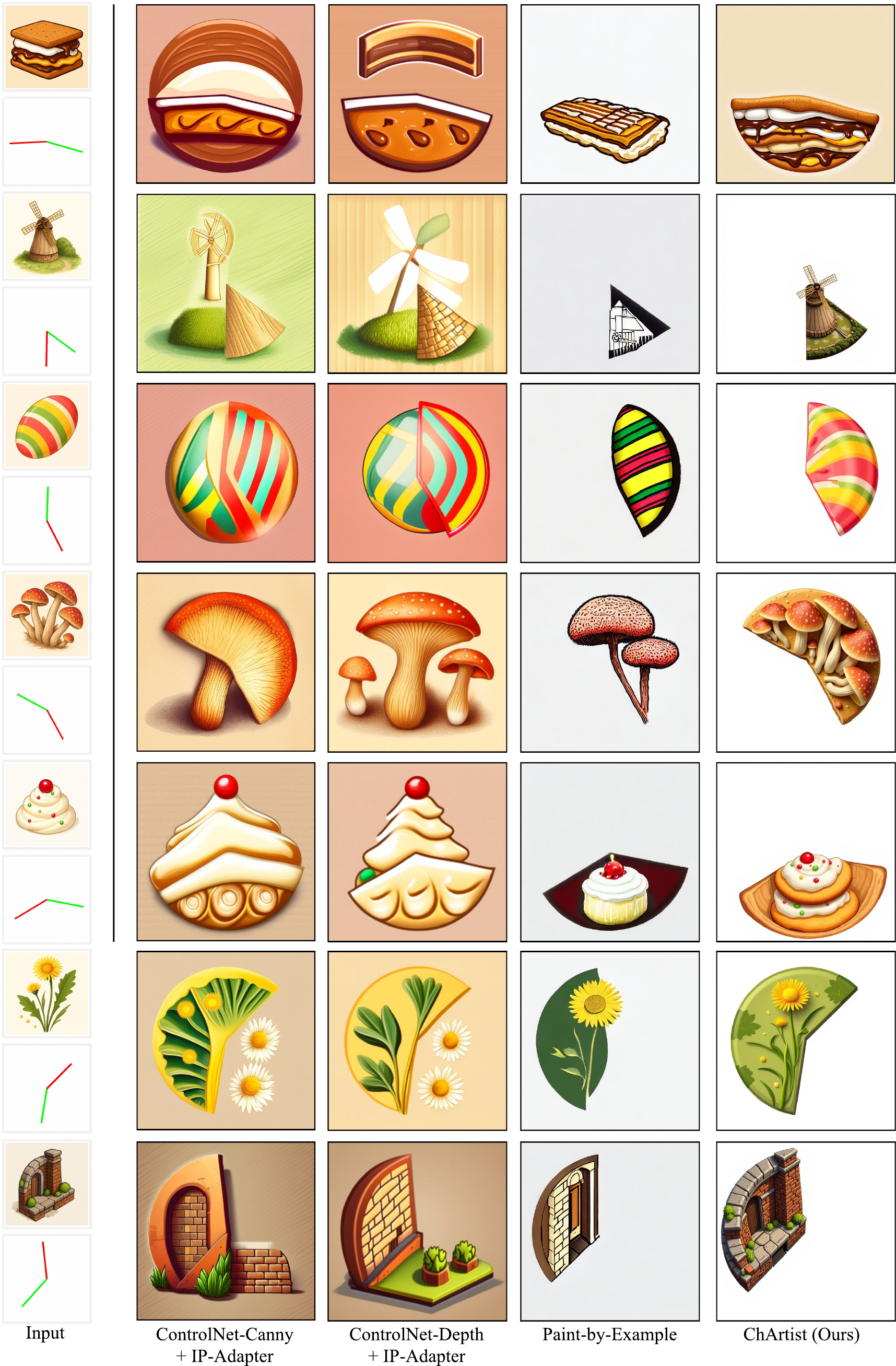} 
  \caption{Dual-control generation pie pictorial results conditioned on both spatial structure and subject reference.}
  \label{fig:supp_sub_pie}
\end{figure*}

\begin{figure*}[t]
  \centering
  \includegraphics[width=0.78\linewidth]{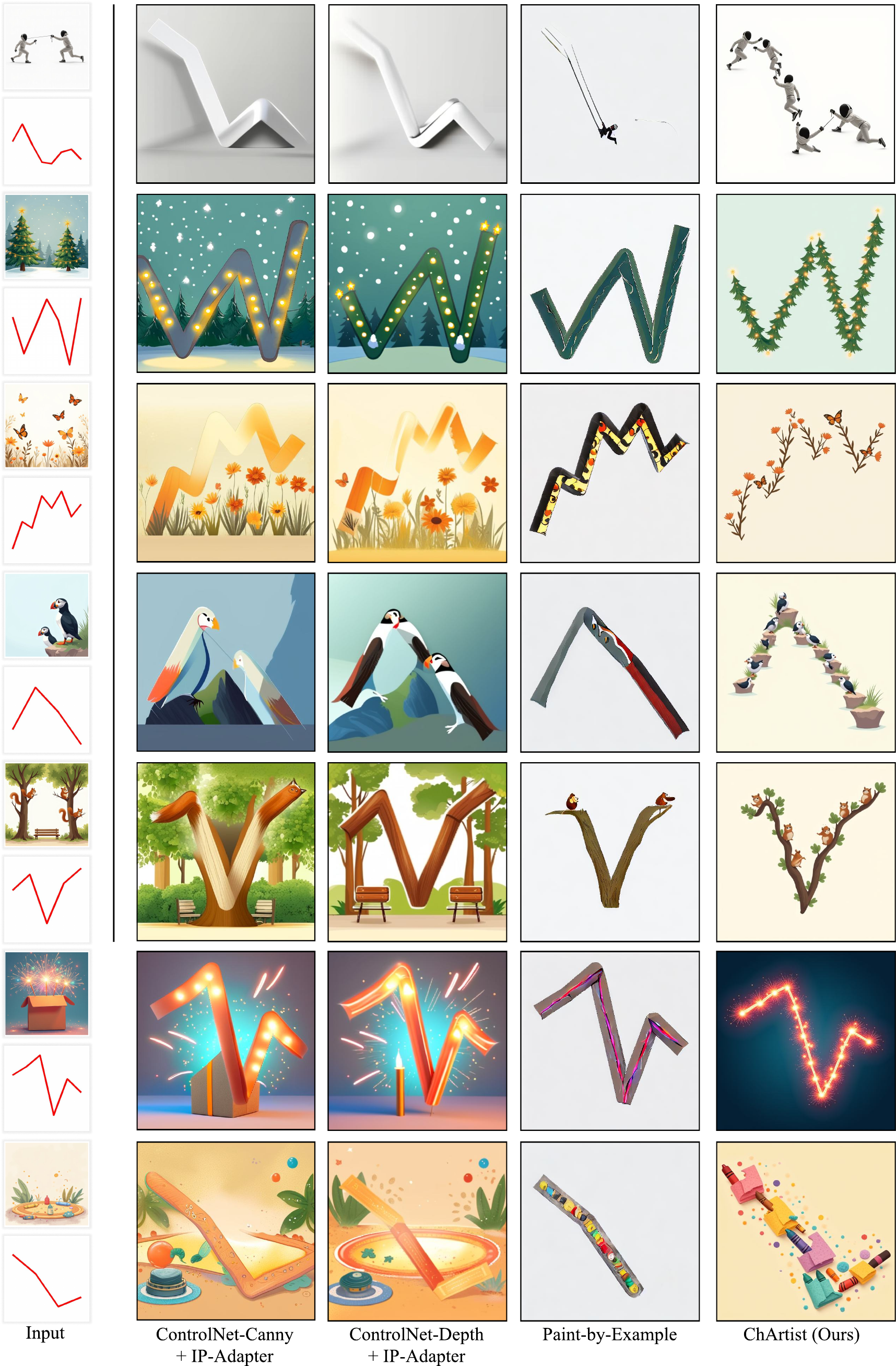} 
  \caption{Dual-control generation line pictorial results conditioned on both spatial structure and subject reference.}
  \label{fig:supp_sub_line}
\end{figure*}

We provide additional qualitative results:

\begin{itemize}
    \item Spatially Aligned Only Task: Fig.~\ref{fig:supp_sk1}, Fig.~\ref{fig:supp_sk2}.

    \item Dual-control generation conditioned on both spatial structure and subject reference: Fig.~\ref{fig:supp_sub_bar}, Fig.~\ref{fig:supp_sub_pie}, and Fig.~\ref{fig:supp_sub_line}.
\end{itemize}



\end{document}